# MACHINE LEARNING VS DEEP LEARNING:
## THE GENERALIZATION PROBLEM

**Yong Yi Bay**
PhD, University of Illinois at Urbana-Champaign
`yongyibay@gmail.com`

**Kathleen A. Yearick**
PhD, University of Illinois at Urbana-Champaign
`kallie.a.yearick@gmail.com`

March 3, 2024

## ABSTRACT

The capacity to generalize beyond the range of training data is a pivotal challenge, often synonymous with a model's utility and robustness. This study investigates the comparative abilities of traditional machine learning (ML) models and deep learning (DL) algorithms in terms of extrapolation – a more challenging aspect of generalization because it requires the model to make inferences about data points that lie outside the domain it has been trained on. We present an empirical analysis where both ML and DL models are trained on an exponentially growing function and then tested on values outside the training domain. The choice of this function allows us to distinctly showcase the divergence in performance when models are required to predict beyond the scope of their training data. Our findings suggest that deep learning models possess inherent capabilities to generalize beyond the training scope, an essential feature for real-world applications where data is often incomplete or extends beyond the observed range. This paper argues for a nuanced understanding of the structural differences between ML and DL models, with an emphasis on the implications for both theoretical research and practical deployment.

## 1 Introduction

Selecting the most appropriate model from a diverse array of algorithms presents a significant challenge within research and industry. Common practice involves exploring a broad spectrum of models and favoring those that demonstrate superior performance. However, this method, as critiqued by Raschka (2018) often resembles a quasi-random, sequential trial and error process rather than a systematic strategy. Such an approach is fundamentally flawed, lacking in both efficiency and theoretical underpinning. It fails to ensure model reliability, robustness, and proper risk management, aspects that are crucial for the deployment of machine learning models in real-world applications (OCC, 2011). Potential performance degradation under unforeseen conditions ultimately limits a model's utility and trustworthiness in critical applications, and may result in financial loss, biased predictions, or other negative consequences.

A principled approach for model selection encompasses considerations that extend beyond mere performance metrics, aiming to address broader implications for model deployment, including reliability and risk. In this context, the concept of generalizability—specifically, a model's ability to perform well on data that it has not encountered during training—emerges as a critical criterion for model selection. This paper argues for the importance of evaluating models based on their capacity for extrapolation. Extrapolation, a demanding aspect of generalization, requires models to make accurate inferences about data points outside their training domain.

Through a detailed examination, we present a test case to compare the extrapolation capabilities of traditional machine learning models against those of deep learning algorithms. This analysis aims to illuminate the differences in how these models generalize beyond their training data, thereby providing insights useful to the decision-making process involved in selecting either machine learning or deep learning models for specific applications. Our findings suggest that a nuanced understanding of a model's generalization and extrapolation capabilities is paramount in making informed selections that align with the desired outcomes and constraints of real-world tasks.

## 2 Methodology

This study aims to compare the abilities of traditional machine learning (ML) models and deep learning (DL) algorithms in terms of extrapolation – a more challenging aspect of generalization because it requires the model to make inferences about data points that lie outside the domain it has been trained on. To achieve this, we construct a scenario that leverages a mathematical function known for its complexity and rapid growth properties. Specifically, we examine the function $f(x) = e^{x^2+x}$, chosen for its non-linearity and the exponential increase that far surpasses that of polynomial functions. This selection ensures a rigorous test of a model's capacity to capture intricate feature interactions — between $x$ and $x^2$ — and to model the function's explosive growth.

### 2.1 Data Generation

We define the domain of $x$ as $\mathcal{D}_x$, to be the closed interval [0, 1]. Within this domain, the target values $y$ are computed using the function $f(x) = e^{x^2+x}$, which presents a significant challenge due to its non-linear growth pattern. To evaluate model performance, we partitioned the dataset into training and testing subsets. The training set, $\mathcal{D}_{\text{train}}$, comprises all $(x, y)$ pairs where $x$ is drawn from the interval [0, 0.7), while the test set, $\mathcal{D}_{\text{test}}$, includes pairs where $x$ falls within [0.7, 1.0]. This partitioning ensures a 70/30 split, which is instrumental in examining the models' capacities to extrapolate beyond the training domain.

$$f(x) = e^{x^2+x}, \text{ where } x \in [0, 1] \tag{1}$$

$$\mathcal{D}_{\text{train}} = \{(x, y) \mid x \in [0, 0.7), y = f(x)\} \tag{2}$$

$$\mathcal{D}_{\text{test}} = \{(x, y) \mid x \in [0.7, 1.0], y = f(x)\} \tag{3}$$

### 2.2 Model Evaluation Criteria

The exponential nature of $f(x)$ suggests that minor deviations from the true function are significantly amplified with increasing $x$, making it an exemplary measure for assessing a model's extrapolation prowess. To quantitatively evaluate model performance, we employ metrics such as $L_1\ error$ − Mean Absolute Error, $L_2\ error$ − Root Mean Squared Error, and $L_\infty\ error$ − Max Absolute Error, across both training and testing sets. These metrics will offer insight into each model's ability to not only fit the training data but also to generalize to unseen data points within the extrapolation domain.

### 2.3 Implementation Details

The models evaluated in this study span a broad spectrum of traditional machine learning (ML) techniques, encompassing both ensemble methods and simpler regression approaches. Specifically, we include XGBoost and LightGBM, which are advanced ensemble learners known for their high performance and efficiency, as well as K-nearest neighbors (KNN) regression and linear regression, to cover a range of algorithmic complexities and learning paradigms. On the deep learning (DL) front, our investigation focuses on a fundamental architecture to illustrate the capabilities of DL models: a fully connected neural network with two hidden layers. This architecture is selected for its straightforward design yet potent ability to capture complex nonlinear relationships, serving as an exemplary representation of deep learning methodologies.

To ensure a fair and rigorous comparison across the board, our methodology incorporates a standardized hyperparameter optimization process tailored to each model's specific needs while maintaining a consistent evaluation framework, including cross-validation within the training dataset for traditional ML models and a validation split approach for DNNs to prevent overfitting and to optimize model architectures. This approach is pivotal in eliminating biases and ensuring that each model is operating at its optimal performance level, thereby providing a level playing field for comparison between traditional machine learning (ML) models and deep learning (DL) models.

For traditional ML models, including XGBoost, LightGBM, K-nearest neighbors regression, and linear regression, we employ a randomized search combined with successive halving as our strategy for hyperparameter tuning. This method begins by evaluating a broad array of hyperparameter configurations using a relatively small resource allocation—such as fewer data points or a limited number of iterations. It then progressively focuses on the most promising configurations by allocating more resources to them in subsequent rounds. This approach not only enhances efficiency by reducing computational waste on less promising model configurations but also ensures a thorough exploration of the hyperparameter space, thereby increasing the likelihood of identifying the optimal settings.

In the realm of DL, specifically for our fully connected neural network with two hidden layers, we adopt the Hyperband approach to hyperparameter optimization. Hyperband is a bandit-based strategy that dynamically allocates resources to a wide range of randomly sampled configurations and rapidly prunes the underperforming models. By iteratively narrowing down the search to configurations that show the most promise, Hyperband effectively balances the exploration-exploitation trade-off, ensuring that computational resources are concentrated on evaluating the most viable neural network architectures.

This parallelism in our approach—successive halving for ML models and Hyperband for DL models—mirrors the underlying principle of efficiently exploring the hyperparameter space while adapting the resource allocation strategy to the specific characteristics of each model type. Such a tailored yet equitable strategy for hyperparameter tuning underlines our commitment to fairness in the comparative analysis. It acknowledges the inherent differences between traditional ML and DL methodologies while striving to optimize each model's architecture within its operational paradigm, ensuring that the comparison of extrapolation capabilities is both equitable and insightful.

## 3 Results

In Table 1, we present a detailed comparison of model performance across three error metrics: $L_1$, $L_2$, and $L_\infty$ norms, both on training and testing data sets. The absolute differences in these metrics, denoted as $|\Delta L_1|$, $|\Delta L_2|$, and $|\Delta L_\infty|$, serve as indicators of each model's extrapolation ability, with lower values suggesting greater model robustness in generalizing beyond the training data.

The Deep Neural Network (DNN) model exhibits markedly lower absolute differences for all three metrics, with $|\Delta L_1|$, $|\Delta L_2|$, and $|\Delta L_\infty|$ all being less than 0.1. This indicates that the performance of the DNN does not degrade substantially when transitioning from the training data to the testing data. Such resilience in extrapolation performance underscores the DNN's ability to capture the underlying data generation process rather than merely fitting to the training data points.

The ensemble methods—XGBoost, LightGBM, and Gradient Boosting—as well as the KNN Regression model, exhibit near-perfect performance on the training set, as indicated by the minimal $L_1$, $L_2$, and $L_\infty$ training errors. However, these models display a significant degradation in performance on the testing set, mirroring the behavior of the linear models. This sharp contrast in train versus test performance, with disparities in errors as evidenced by the $|\Delta L_1|$, $|\Delta L_2|$, and $|\Delta L_\infty|$ metrics, underscores their limited extrapolation capabilities when applied to data beyond their training domain.

Table 1: Train and test set model performance

| Models | $L_1$ | | | $L_2$ | | | $L_\infty$ | | |
|---|---|---|---|---|---|---|---|---|---|
| | Train | Test | $|\Delta L_1|$ | Train | Test | $|\Delta L_2|$ | Train | Test | $|\Delta L_\infty|$ |
| **Deep Neural Network** | 4.3E-03 | 5.3E-02 | **4.9E-02** | 5.5E-03 | 5.9E-02 | **5.3E-02** | 2.3E-02 | 8.9E-02 | **6.6E-02** |
| XGBoost | 4.1E-03 | 1.7E+00 | 1.7E+00 | 5.2E-03 | 2.1E+00 | 2.1E+00 | 1.8E-02 | 4.1E+00 | 4.1E+00 |
| LightGBM | 4.2E-03 | 1.8E+00 | 1.8E+00 | 9.9E-03 | 2.1E+00 | 2.1E+00 | 8.8E-02 | 4.2E+00 | 4.1E+00 |
| Gradient Boosting | 2.4E-03 | 1.7E+00 | 1.7E+00 | 3.2E-03 | 2.1E+00 | 2.1E+00 | 1.2E-02 | 4.1E+00 | 4.1E+00 |
| Random Forest | 7.9E-04 | 1.7E+00 | 1.7E+00 | 1.0E-03 | 2.1E+00 | 2.1E+00 | 5.4E-03 | 4.1E+00 | 4.1E+00 |
| KNN Regression | 2.8E-05 | 1.7E+00 | 1.7E+00 | 3.2E-04 | 2.1E+00 | 2.1E+00 | 5.2E-03 | 4.1E+00 | 4.1E+00 |
| Linear Regression | 1.4E-01 | 1.7E+00 | 1.6E+00 | 1.6E-01 | 1.9E+00 | 1.8E+00 | 4.4E-01 | 3.6E+00 | 3.2E+00 |
| Huber Regression | 1.3E-01 | 1.8E+00 | 1.7E+00 | 1.6E-01 | 2.0E+00 | 1.8E+00 | 5.1E-01 | 3.7E+00 | 3.2E+00 |
| Ridge Regression | 1.4E-01 | 1.7E+00 | 1.6E+00 | 1.6E-01 | 1.9E+00 | 1.8E+00 | 4.4E-01 | 3.6E+00 | 3.2E+00 |
| Bayesian Ridge Regression | 1.4E-01 | 1.7E+00 | 1.6E+00 | 1.6E-01 | 1.9E+00 | 1.8E+00 | 4.4E-01 | 3.6E+00 | 3.2E+00 |

The training phase, denoted by the white background in Figure 1, shows DNN, Ensemble, and KNN models all closely approximating the true function, indicated by the solid pink line. This is expected as the models are optimized to reduce error on the training data. The DNN, Ensemble, and KNN models have learned the pattern within the range of training data with high fidelity.

Transitioning into the testing phase, which is highlighted by a red background corresponding to $x \in [0.7, 1.0]$ a pronounced divergence in model performance emerges. A plateauing of predictions just beyond the demarcation at $x = 0.7$ is observed in the Ensemble methods—XGBoost, LightGBM, and Gradient Boosting—and the KNN Regression model. This plateau indicates a critical limitation in these models: their inability to extrapolate the non-linear function's complexity outside of the scope of the training data. Rather than continuing the true function's trend, these models default to projecting a horizontal extrapolation of the last learned value, resulting in a constant output.

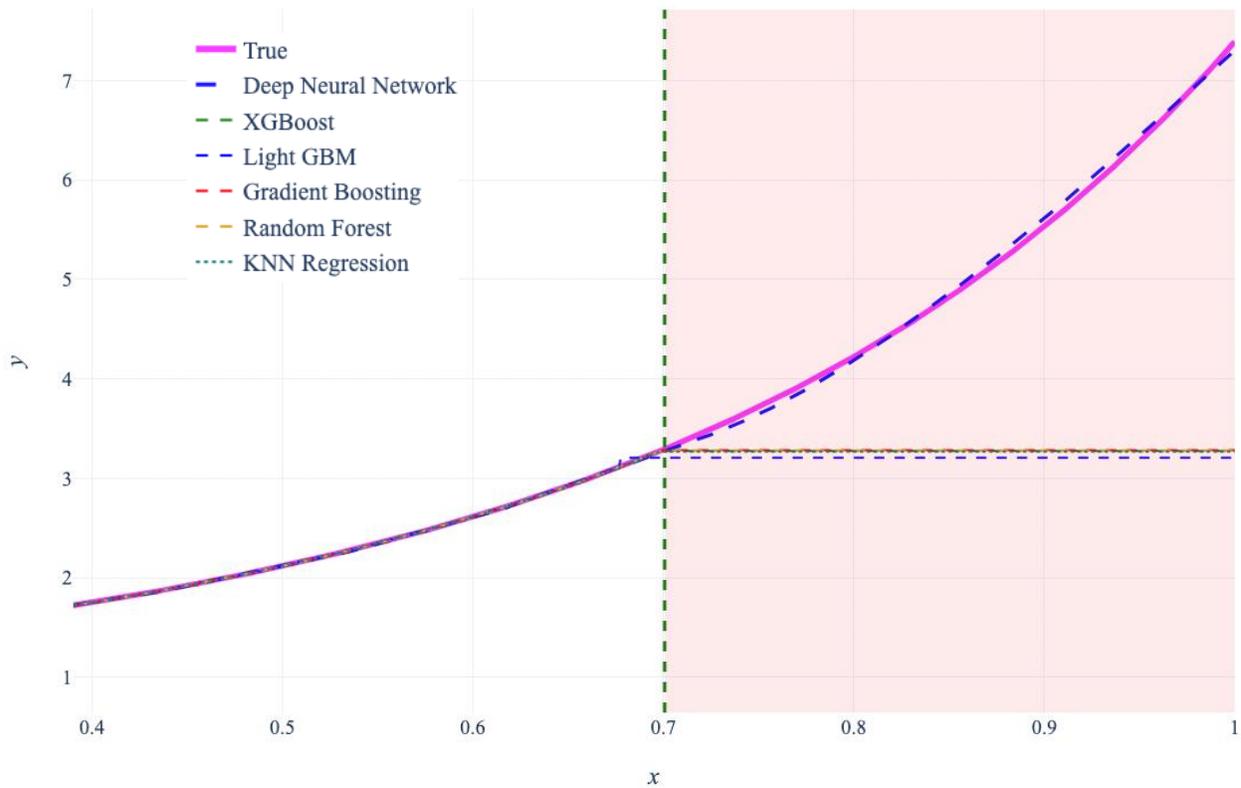

Figure 1: DNN vs Ensemble and KNN models across the interval $x \in [0.4, 1.0]$ with the white and red shaded backgrounds representing the training and testing datasets, respectively.

For the linear models, Figure 2 elucidates the underlying factors contributing to the suboptimal performance of linear models as observed in Table 1. Within the depicted interval $x \in [0.4, 1.0]$, the linear models—Linear Regression, Huber Regression, Ridge Regression, and Bayesian Ridge Regression—struggle to conform to the non-linear profile of the true function, represented by the purple dashed line. This limitation stems from their inherent design to capture linear relationships, which compels them to establish the best-fitting straight line that minimizes error across the training data, shaded in white.

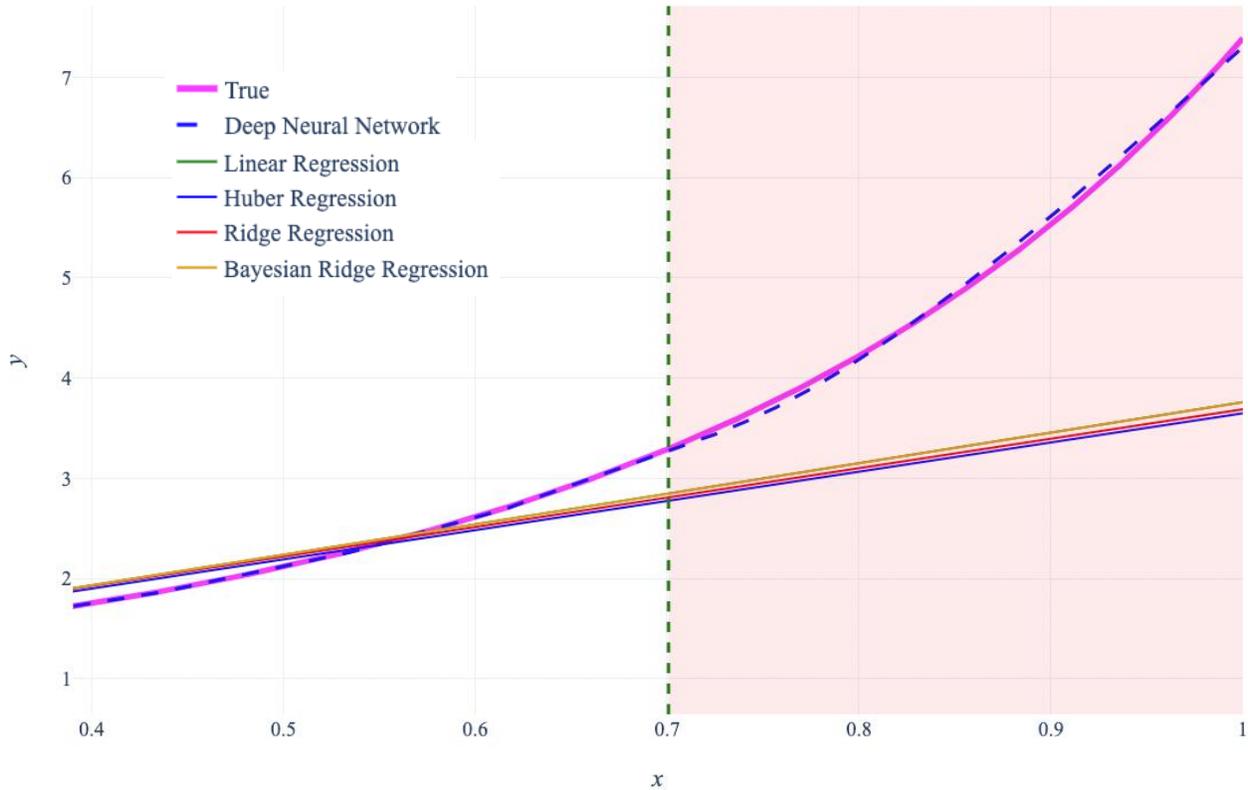

Figure 2: DNN vs linear models across the interval $x \in [0.4, 1.0]$ with the white and red shaded backgrounds representing the training and testing datasets, respectively.

The DNN's superior performance can be attributed to its deep architecture and non-linear activation functions, which provide the flexibility and capacity necessary to learn and generalize complex functions. This is consistent with the literature that acknowledges the strength of neural networks in modeling complex, high-dimensional mappings due to their hierarchical structure and distributed representation capabilities.

## 4 Conclusion and Broader Impact

Our study provides a clear indication of the superior ability of deep learning (DL) to extrapolate beyond the confines of training data, a demonstration of its remarkable generalization capabilities. When contrasted with the array of traditional machine learning (ML) models evaluated, which tend to exhibit behavior akin to data memorization, DL stands out. Traditional models, in this analysis, appear to operate comparably to a look-up table, proficient in regurgitating observed information yet faltering in the prediction of unseen data.

However, these findings should not be misconstrued as an endorsement for the wholesale replacement of ML with DL approaches. It is crucial to recognize that the assortment of models explored in this research is not exhaustive. Moreover, DL models come with their own set of prerequisites, including—but not limited to—substantial data requirements, computational resources, and a longer training period, which may not be feasible or necessary for all applications. In scenarios where there is an abundance of training data and there is strong assurance that future data will mirror the training set, machine learning remains a valuable resource.

In the broader context, the implications of this research are manifold. The discernible prowess of DL in extrapolation is particularly relevant in scenarios where the cost of misprediction is high, such as medical diagnosis, financial forecasting, and autonomous systems. These fields can greatly benefit from models that can accurately infer outcomes in situations that deviate from prior experiences. Conversely, when interpretability and resource constraints are paramount, traditional ML models may still hold their ground as the more appropriate choice.

In conclusion, our investigation underscores the need for a balanced and informed approach for model selection. It advocates for a careful consideration of the trade-offs between the robust generalization of DL and the practical advantages of traditional ML models. As we continue to advance in our understanding and development of predictive models, it becomes increasingly critical to tailor our choices to the specific requirements and limitations of the task at hand.

# Appendix A: Model Hyperparameters

### A.1 Deep Neural Network Hyperparameters

Parameters tuned in our experiment:

| Parameters | Value |
|---|---|
| 'units' in first hidden layer | 512 |
| 'units' in second hidden layer | 448 |
| 'learning_rate' | 0.01 |

*Note: Other parameters that were chosen but not tuned are 'optimizer': 'Adam', 'loss': 'mse', 'activation': 'relu'. Parameters that are not listed use default values as per TensorFlow documentation for v2.15*

### A.2 XGBoost Hyperparameters

Parameters tuned in our experiment:

| Parameters | Value |
|---|---|
| 'n_estimators' | 157 |
| 'max_depth' | 3 |
| 'learning_rate' | 0.20 |
| 'subsample' | 0.73 |
| 'colsample_bytree' | 0.88 |
| 'min_child_weight' | 0.1 |

*Note: Parameters that are not listed use default values as per XGBoost documentation for v2.0.3*

### A.3 LightGBM Hyperparameters

Parameters tuned in our experiment:

| Parameters | Value |
|---|---|
| 'n_estimators' | 279 |
| 'max_depth' | 8 |
| 'learning_rate' | 0.17 |
| 'subsample' | 0.83 |
| 'colsample_bytree' | 0.75 |
| 'min_child_weight' | 0.01 |

*Note: Parameters that are not listed use default values as per LightGBM documentation for v4.3.0.*

### A.4 Gradient Boosting Regressor Hyperparameters

Parameters tuned in our experiment:

| Parameters | Value |
|---|---|
| 'n_estimators' | 259 |
| 'max_depth' | 3 |
| 'learning_rate' | 0.10 |
| 'min_samples_split' | 8 |

*Note: Parameters that are not listed use default values as per scikit-learn documentation for v1.4.1.*

### A.5 Random Forest Regressor Hyperparameters

Parameters tuned in our experiment:

| Parameters | Value |
|---|---|
| 'n_estimators' | 195 |
| 'max_depth' | 9 |
| 'min_samples_split' | 2 |
| 'min_samples_leaf' | 1 |

*Note: Parameters that are not listed use default values as per scikit-learn documentation for v1.4.1.*

### A.6 K-Nearest Neighbors Regressor Hyperparameters

Parameters tuned in our experiment:

| Parameters | Value |
|---|---|
| 'n_neighbors' | 2 |
| 'weights' | 'distance' |
| 'algorithm' | 'ball-tree' |

*Note: Parameters that are not listed use default values as per scikit-learn documentation for v1.4.1.*

### A.7 Linear Regression Hyperparameters

Linear regression does not have hyperparameters to tune, as it is a straightforward model without parameters that affect the fitting process.

### A.8 Huber Regressor Hyperparameters

Parameters tuned in our experiment:

| Parameters | Value |
|---|---|
| 'epsilon' | 1.35 |
| 'alpha' | 0.1 |

*Note: Parameters that are not listed use default values as per scikit-learn documentation for v1.4.1.*

**A.9 Ridge Regressor Hyperparameters**

Parameters tuned in our experiment:

| Parameters | Value |
|---|---|
| 'alpha' | 0.1 |

*Note: Parameters that are not listed use default values as per scikit-learn documentation for v1.4.1.*

**A.10 Bayesian Ridge Regressor Hyperparameters**

Parameters tuned in our experiment:

| Parameters | Value |
|---|---|
| 'max_iter' | 100 |
| 'alpha_1' | 1e-7 |
| 'alpha_2' | 1e-5 |
| 'lambda_1' | 1e-5 |
| 'lambda_2' | 1e-7 |

*Note: Parameters that are not listed use default values as per scikit-learn documentation for v1.4.1.*